\documentclass[conference,letter,10pt,final]{llncs}
\usepackage[utf8]{inputenc}
\usepackage[T1]{fontenc}
\usepackage{graphicx}
\usepackage{grffile}
\usepackage{longtable}
\usepackage{wrapfig}
\usepackage{rotating}
\usepackage[normalem]{ulem}
\usepackage{amsmath}
\usepackage{textcomp}
\usepackage{amssymb}
\usepackage{capt-of}
\usepackage{hyperref}
\usepackage[utf8]{inputenc}
\usepackage[T1]{fontenc}
\date{}
\title{Solving Fashion Recommendation - The Farfetch Competition}
\hypersetup{
 pdfauthor={Aditya Jain},
 pdftitle={Solving Fashion Recommendation - The Farfetch Competition},
 pdfkeywords={},
 pdfsubject={},
 pdfcreator={Emacs 28.0.50 (Org mode 9.5)}, 
 pdflang={British}}
\begin{document}

\title{Solving Fashion Recommendation - The Farfetch Challenge}
%\titlerunning{}
%\toctitle{}
%\subtitle{}

\author{Manish Pathak\inst{1} \and Aditya Jain\inst{2}}
%\authorrunning{}
%\tocauthor{}
\institute{
MiQ Digital, Bengaluru, India \\
\email{manishpathak@miqdigital.com}
\and
MiQ Digital, Bengaluru, India \\
\email{aditya@miqdigital.com}}

% \thanks can be used anywhere in author, institute and title

\maketitle

\begin{abstract}

Recommendation engines are integral to the modern e-commerce experience, both
for the seller and the end user. Accurate recommendations lead to higher revenue
and better user experience.
In this paper, we are presenting our solution to ECML PKDD Farfetch
Fashion Recommendation Challenge \cite{farfetch2021challenge}. The goal of this challenge is to maximize the
chances of a click when the users are presented with set of fashion items.
We have approached this problem as a binary classification problem. Our winning
solution utilizes Catboost as the classifier and Bayesian Optimization for hyper
parameter tuning. Our baseline model achieved MRR of 0.5153 on the validation set.
Bayesian optimization of hyper parameters improved the MRR to 0.5240 on the
validation set. Our final submission on the test set achieved a MRR of 0.5257.

\keywords\{Fashion Recommendation \and Catboost \and Hyper parameter tuning \and
Classification\}
\end{abstract}

\section{Introduction}
\label{sec.intro}
In these times of ubiquitous internet access and ever shortening attention span
it is imperative to capture your users' attention before they hop to a different
website or an app or worse, to your competitors' website. The friction-less
nature of web mandates the content to align to your users' liking. Whether it be
a blog, social media platform, e-commerce store, search engine or any other form
of information aggregator, usage of a recommendation systems is imperative.

Industries like Social Media and Digital Advertisement see extensive use of such
systems to reduce cost and increase their Click Through Rate(CTR). This
competition too is structured along similar lines of predicting clicks, and
recommending only those products to the users that have a high probability of
being clicked.

Our goal in this competition was to use a technique that can work with modest
resources of a personal laptop as opposed to a distributed cluster in the cloud
or a dedicated deep learning workstation. We traded
resources with time. Our final model took approximately 35 hours to train including hyper
parameter optimization. In this paper, we will be discussing the problem
statement, our approach to modelling, and results of our approach.

We have divided this paper into 5 sections. Section \ref{sec.data} describes the
problem statement and the provided datasets in detail. Section \ref{sec.model}
deals with data preparation, modelling, training, and tuning. We discuss the
results in section \ref{sec.result}.

\section{Dataset and Problem Statement}
\label{sec.data}
Before we dive into the dataset, we would like to explain the
terminology used in the competition. An \emph{impression} is an unordered set of six
products that were shown to a user. If the user clicks on any of the recommended
products, it is called a \emph{click}. The dataset contains information about these
impressions along with contextual information. Following are the features
provided in the impressions dataset:
\begin{itemize}
\item \emph{query\_id}: Unique ID of impression
\item \emph{user\_id}: Unique ID of the user this impression was shown to
\item \emph{session\_id}: A user can have many sessions. This column contains the session
id corresponding to a particular impression
\item \emph{product\_id}: Identifier or product shown
\item \emph{page\_type}: Type of page where the recommendation was shown
\item \emph{previous\_page\_type}: Immediately previous page that the user visited
\item \emph{device\_category}: Type of device
\item \emph{device\_platform}: Platform running on the device
\item \emph{user\_tier}: User's level in Farfetch's reward program
\item \emph{user\_country}: User's country
\item \emph{context\_type}: Comma separated list of types of contexts
\item \emph{context\_value}: Corresponding identifiers of contexts from \emph{context\_type} field
\item \emph{product\_price}: Normalized price of the product
\item \emph{week}: Normalized Week number of impression
\item \emph{week\_day}: Day of week when the impression was shown
\item \emph{is\_click}: This is the target variable. Contains 1 if the user clicked on the
product, 0 otherwise.
\end{itemize}
The impression dataset contains rows corresponding to each product
recommendation shown.
The attribute dataset contains information about individual products. Following
are the product attributes available in this dataset:
\begin{itemize}
\item \emph{product\_id}: Identifier of product, used to do an inner join with impressions dataset
\item \emph{gender}: Product's target gender
\item \emph{main\_colour}: Principal colour of the product
\item \emph{second\_colour}: The second most predominant colour of the product
\item \emph{season}: Fashion season of the product
\item \emph{collection}: Fashion collection identifier
\item \emph{category\_id\_l1}: Top level product category
\item \emph{category\_id\_l2}: Second level of product category
\item \emph{category\_id\_l3}: Third level of product category
\item \emph{brand\_id}: Product brand ID
\item \emph{season\_year}: Year of the fashion season
\item \emph{start\_online\_date}: Number of days the product has been online with respect
to a predefined reference date
\item \emph{attribute\_values}: Comma separated list of miscellaneous product attributes
\item \emph{material\_values}: Comma separated list of product's material composition
\end{itemize}

The dataset contained 3,507,990 anonymized events captured over a duration of 2 months by Farfetch's
internal recommender system
The \emph{context\_value} field
contained 189,571 unique values.  Similarly, the \emph{session\_id} field contained
317,426 unique values. The cardinality of other features was lower. The
\emph{page\_type} column had 5 unique values whereas \emph{previous\_page\_type} field had 23
unique values.

The test dataset had a reasonable overlap in terms of unique values in these
fields. 23.82\% of \emph{session\_id} values in the training data occured in the test
dataset.  Similarly, 23.88\% of \emph{context\_value} values occured in the test set.
The \emph{user\_cuuntry}, \emph{page\_type}, and \emph{previous\_page\_type} fields had 88.26\%,
80\%, and 95.65\% unique values from the training set occur in the test
dataset.

The target of this modelling is to predict a set of 6 products having the
maximum likelihood of being clicked by the user based on the provided impression
context. The recommended set of 6 products need to be ordered from most likely
to least likely of being clicked. The evaluation metric is Mean Reciprocal Rank
which we will discuss in section \ref{sec.metric}.

\section{Modelling Approach}
\label{sec.model}
\subsection{Modelling Overview}
\label{sec:org3b62550}
As discussed in the previous section, the problem of ranking products based on
the preference of the user running a query/impression in a given session is
essentially a type of Recommender Systems problem, more specifically, an
Implicit Recommender Systems problem \cite{hu2008}. Instead of building a recommender
system to rank the products in a query, we started approaching the problem from
a binary classification viewpoint because we had the information about labelled
interactions of each user with a product in the binary target feature
\(is\_click\). The idea was to rank the products in a query based on the sorted
click probabilities from highest to lowest generated by the binary classifier.
The feature set that we kept to build the baseline model was all the features
that were present in the validation set and test set.
\subsection{Data Preprocessing}
\label{sec:orgb87f5db}
There were a total of 15 hashed feature vectors (\(X\)) present in the training set.
Out of these 15 feature vectors, 14 feature vectors were categorical in nature
and 1 feature vector was a continuous feature. The target feature \(is\_click\) (\(Y\))
was also categorical (binary) in nature.

Since the categorical feature values were all hashed with random signed
integers/strings, we started treating them with label-encoding. Suppose a
categorical feature vector \(v\) has cardinality of \(m\), then label-encoding will
change the \(i^{th}\) feature value \(v_i\) such that

\begin{equation}
v_i = i, 1\le i \le m ; v_i \in v
\end{equation}

We have applied same mapping for both train and test datasets. For cases
where there were other new categories present in a feature vector in validation
and test sets, the encoding will label new numbers to those categories.

As an initial step to treat the missing values in the training set we replaced
the missing values in categorical features with an arbitrary integer -999 and
missing values in continuous features as the mean of that feature.
\subsection{Choosing a Classifier}
\label{sec:org7f540c7}
As seen in section \ref{sec.data}, the training data set was tabular in nature and
had more than 3 million rows. The colums \emph{session\_id}, and \emph{context\_value} had
extremely high cardinality. Consequently, the capability of handling high
cardinality features and large dataset with limited resources motivated our
choice of algorithm - Catboost \cite{prokhorenkova2019}. Catboost is a new
boosting framework that often outperforms other boosting implementations in
terms of quality of prediction especially in case of tabular heterogeneous data
\cite{hancock2020,zotero-182}. Being a Gradient Boosted Tree method, Catboost also
has the added benefit of providing feature importance which helps in
interpreting the result.

The classical approach of handling categorical data is one-hot encoding \cite{cohen2014a,chapelle2015,micci-barreca2001a}.
But the problem with one-hot encoding is that it creates high-dimensional
feature vectors and this exacerbated in case of features with high cardinality.
Other methods like target encoding \cite{micci-barreca2001a} may also lead to target leakage as discussed in \cite{prokhorenkova2019}.
Other classifiers such as XGBoost consumes more memory than Catboost \cite{hancock2020}. Another reason to choose Catboost, over other bagging-based classifier
such as RandomForest, was its ability to handle categorical feature vectors,
which is central to its design. Catboost performs Ordered Target Encoding for
categorical columns by performing random permutations of the dataset and then
target encoding each example using only the objects that are placed before the
current object. This ordered target encoding solves the problem of target
leakage and is shown to produce better results as compared to other encoding
methods \cite{prokhorenkova2019}. It can also create new categorical features combining the existing
ones and has a method to handle encodings for new categories in the test set that
haven't appeared in the training set using priors \cite{prokhorenkova2019}. This capability of Catboost allowed us to avoid expensive feature engineering.

\subsection{Evaluation Metric}
\label{sec.metric}
The goal of this competition was to rank the products by their likelihood of being clicked, hence we
chose to keep track of the Logloss of the predictions. Mathematically, Logloss
for a single training example is defined as:

\begin{equation}
L_{log}(y,p) = -(y\log(p) + (1-y)\log(1-p))
\end{equation}

where p is the probability of the class given by the learned model and y is the
actual label of the class.

It is important to note that since the final goal is to rank the products, only the relative probabilities matter instead of the actual magnitude of
the probabilities and because of this the precision-based metrics are not very
important to evaluate the model in this case.

The final evaluation metric in the competition was Mean Reciprocal
Rank (MRR). When considering a set of \(n\) queries/impressions, this metric can be
expressed as:

\begin{equation}
MRR = \frac{1}{\left|{n}\right|}\sum_{i=1}^{\left|{n}\right|}\frac{1}{rank_i}
\end{equation}

Where, for the \(i^{th}\) impression in the dataset, \(rank_i\) is the rank of the first
correct prediction.

\subsection{Model Tuning}
\label{sec.tuning}
In order to improve the Logloss, we decided to carry out hyper-parameter tuning
of our baseline Catboost model. General methods to do hyper-parameter
optimizations include Grid-Search and Random-Search but in this case in-order to
tune the hyper-parameters we used Sequential model-based optimization also known
as Bayesian Optimization \cite{shahriari2016}. Bayesian optimization is an efficient method to
find global maximiser/minimiser of an unknown objective function \(f()\)

\begin{equation}
x^* = argmax\ f(x), x \in \chi
\end{equation}

Where \(\chi\) is some space of interest \cite{sanders2019}. Bayesian optimization performs hyperparameter
search in an informed manner i.e. it keeps track of objective score,that is to
be minimized or maximised, of the objective function for a set of
hyperparameters in each trial and then selects the next best set of
hyperparameters based on the past trials' results. The time spent in
selecting hyperparameters in an informed way is inconsequential compared to the
time spent in searching for hyperparameters over a random space like that in
Grid-Search or Random Search and thus Bayesian methods can find best
hyperparameters in relatively fewer trials. In order to
perform this optimization we used the python library Hyperopt \cite{bergstra2013}.
The Catboost parameters we chose to tune were \(learning\_rate\), which controls
the gradient step in minimization methods such as Gradient descent, and the
\(l2\_leaf\_reg\), which is the coefficient of L2 regularization term of the cost
function. The search space for these parameters was configured as follows:
\begin{enumerate}
\item \emph{learning\_rate}: Search over a uniform distribution between 1e-3 to 5e-1.
\item \emph{l2\_leaf\_reg}: Quantized log uniform search space \emph{qlognormal} \cite{zotero-184} with low=0, high=2, and q=1
\end{enumerate}

The final values of these hyperparameters after tuning were:
\begin{enumerate}
\item \emph{learning\_rate}: 0.16610
\item \emph{l2\_leaf\_reg}: 2
\end{enumerate}

The algorithm that we choose for searching was Tree-structured Parzen Estimator Approach \cite{bergstra}

\section{Results}
\label{sec.result}
We trained the baseline model with stratified K-Fold Cross Validation on the
entire training data to make the predictions more robust to high variance for N
number of iterations. Every iteration updates \(N\) models. We evaluate each model
on its own validation dataset on every iteration. This produces \(N\) metric
values on each iteration \(K\).

\begin{figure}[htbp]
\centering
\includegraphics[width=.9\linewidth]{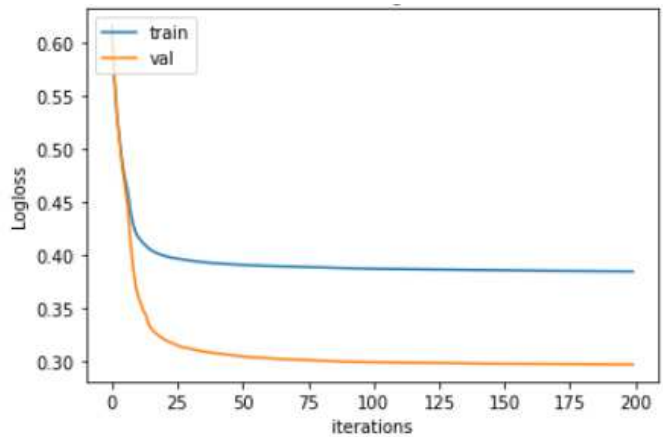}
\caption{Training and Validation logloss during hyper parameter optimization \label{img.loss}}
\end{figure}

\begin{figure}[htbp]
\centering
\includegraphics[width=.9\linewidth]{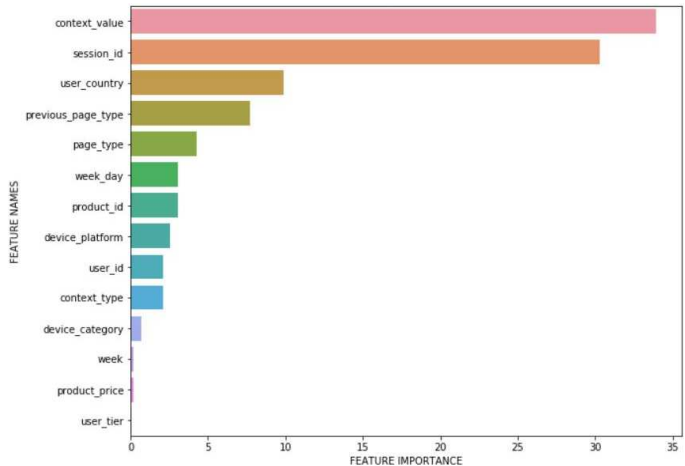}
\caption{Importance of various features as calculated by the model \label{img.fi}}
\end{figure}

Using this model, we predicted the probability of click on the products present
in the validation set. We ranked the probabilities from 1 to 6 by grouping over
the \(query\_id\) column. The baseline model gave an MRR of 0.51526 on the
validation-set and a mean training logloss of 0.393526. Mean validation logloss
for baseline model stood at 0.309565.

The next stage of training was hyper parameter tuning as discussed in
\ref{sec.tuning}. We saw a substantial improvement in MRR. Validation set MRR
stood at 0.52396 whereas final test set MRR stood at 0.52570 after hyper
parameter tuning. We also saw substantial improvement in the training and
validation logloss numbers which stood at 0.385014 and 0.297610 respectively. The
figure \ref{img.loss} shows the progression of logloss on training and validation
set during this phase. The training and validation logloss followed similar
trend throughout the training process which indicates that our model did not overfit.

Figure \ref{img.fi} presents the feature importance graph that was obtained after
the training. The feature \emph{context\_value} turned out to be the most important
feature. This is not surprising and is well known in digital advertising that
publisher features which includes \emph{URL} are very important
\cite{pathak2020,chapelle2015}. The \emph{session\_id} feature turned out to be second most important as per figure
\ref{img.fi} surpassing the \emph{product\_id} feature which stands for the current product
being viewed by the user.

\section{Conclusion}
\label{sec.conclusion}
There are 3 points that we would like to highlight in the conclusion:
\begin{enumerate}
\item Modern Algorithms like catboost perform really well without dedicated
feature engineering even in cases of very high cardinality features.
\item Bayesian Hyper Parameter Optimization is an extremely effective way to
improve model performance.
\item It is possible to build and train recommendation systems with modest resources.
Distributed systems are not always necessary.

We would like to emphasise the last point. Data driven recommendations are
pervasive. Business not leveraging such a system due to cost concerns are not
utilizing data effectively. Cost of building a data driven recommendation
system has never been lower and more democratic. Therefore, cost should not
be seen as a barrier to start implementing such a system.
\end{enumerate}
\bibliographystyle{splncs04}
\bibliography{refs}
\end{document}